\documentclass{article} 
\usepackage{iclr2027_conference,times}

\usepackage{amsmath,amsfonts,bm}

\def\eqref#1{equation~\ref{#1}}

\def\1{\bm{1}}

\DeclareMathAlphabet{\mathsfit}{\encodingdefault}{\sfdefault}{m}{sl}
\SetMathAlphabet{\mathsfit}{bold}{\encodingdefault}{\sfdefault}{bx}{n}

\DeclareMathOperator*{\argmax}{arg\,max}

\usepackage{enumitem}
\usepackage{graphicx}
\usepackage{subcaption}
\usepackage{placeins}
\usepackage{etoc}
\usepackage{fvextra}
\usepackage{hyperref}
\usepackage{url}

\usepackage{relsize}
\usepackage[ruled,linesnumbered,algo2e]{algorithm2e}
\usepackage{xcolor}
\usepackage{array}
\usepackage{booktabs}
\usepackage{longtable}

\title{\vspace{-15pt}Anticipating the Consequences of Curriculum Decisions with Large Language Models}

\author{
Octavio Pappalardo\thanks{Correspondence to: \texttt{octaviopappalardo@gmail.com}. \\ \hspace*{1.6em} Code: \url{https://github.com/Octavio-Pappalardo/curriculum-foresight}} \\
{\normalfont University College London}
\And
Nathan Herr \\
{\normalfont University College London}
\And
Tim Rocktäschel \\
{\normalfont University College London}
}

\iclrfinalcopy 
\begin{document}

\maketitle
\lhead{}
\renewcommand{\headrulewidth}{0pt}
\vspace{-4pt}
\begin{abstract}
\vspace{-4pt}
Automatic curriculum learning can improve the effectiveness of reinforcement learning by selecting the training experiences presented to the agent over time. Predicting the consequences of such decisions can, however, be difficult. We analyze automatic curriculum learning as a sequential decision-making problem, highlighting a gap between the quantities that determine the value of curriculum decisions and the information captured by local learning signals commonly used to guide them. We then investigate whether Large Language Models (LLMs) can exploit richer information about the learning problem to better anticipate the consequences of curriculum decisions. We introduce a method that combines online learning-progress estimates with LLM-informed estimates of (i) the potential downstream benefits of learning on each task and (ii) whether direct training on a task is currently likely to produce progress. We evaluate the approach on a custom benchmark of 256 textual goals in Craftax under different curriculum objectives. We observe the strongest gains when optimizing for individual target tasks. When optimizing across the full task set, the benefits vary across learners with different mechanisms for cross-task transfer, ranging from modest improvements in learning speed to larger gains that persist through the end of training.
\end{abstract}

\etocdepthtag.toc{main}

\vspace{-4pt}
\section{Introduction}\label{sec:Introduction}

Reinforcement learning tasks can be difficult to learn through direct training, particularly when rewards are sparse or delayed. Curriculum learning addresses this by shaping the sequence of situations on which an agent trains, a principle that has played an important role in several influential deep reinforcement learning systems \citep{akkaya2019solving,vinyals2019grandmaster,hubert2026olympiad}. Choosing these training experiences, however, is itself a complex sequential decision-making problem \citep{narvekar2017autonomous}. Evaluating the value of training on a task requires anticipating its immediate impact on the agent's performance across tasks, as well as its impact on the effectiveness of future training. These consequences are difficult to predict, and commonly used curriculum signals such as estimates of task competence or learning progress are limited for this purpose \citep{matiisen2019teacher}.

In this work, we analyze automatic curriculum learning explicitly from this sequential decision-making perspective. Using learning progress as a concrete lens, we relate local task-selection criteria to optimal curriculum decisions and make explicit how unaccounted downstream effects of training and limitations in estimating local progress can lead to suboptimal decisions. We then investigate whether Large Language Models (LLMs) can help account for these factors by exploiting richer information about the learning problem. We introduce a method in which an LLM, given descriptions of the tasks, learner, and training setup, provides judgments that help estimate quantities affecting the value of curriculum decisions. Concretely, we augment learning-progress-based prioritization with estimates of the potential downstream benefits of training on a task and of whether direct training on a low-performance task is likely to produce progress. We evaluate the approach on a custom benchmark of 256 textual goals in Craftax and find that it can materially improve learning outcomes. When optimizing for individual target tasks, the method makes substantial progress on targets where learning-progress-based prioritization makes little or none. When optimizing across the full task set, gains are modest with a learner restricted to primitive actions but larger for a learner that can invoke learned behaviors as subroutines.

\section{Related work}\label{sec:Related Work}
One approach to automatic curriculum learning is to directly optimize a curriculum-level objective defined over the overall learning process \citep{foglino2019optimization}. For instance, \citet{narvekar2018learning} formulate curriculum selection as a higher-level MDP and use reinforcement learning to learn a curriculum policy from repeated executions of complete curriculum trajectories. Such approaches can in principle account for the full consequences of curriculum decisions, but evaluating candidate curricula entails repeatedly training the underlying learner, making direct optimization highly sample and computationally expensive \citep{narvekar2018learning,foglino2019curriculum,foglino2019optimization}.

In practice, most research on automatic curriculum learning has focused on methods that construct the curriculum within a single learner training run, using information available at the start of training or captured online to construct proxies for the utility of training on different tasks \citep{portelas2020automatic}. One family of approaches uses measures of diversity or novelty to guide task selection, favoring tasks that are underrepresented in the learner's experience \citep{pong2019skew,pitis2020maximum}. Related approaches use measures of surprise, such as prediction error or model disagreement, to prioritize situations that remain unfamiliar to the learner \citep{pathak2017curiosity,pathak2019self}. Another widely used principle is to favor tasks of intermediate difficulty relative to the learner's current capabilities, as tasks that are reliably solved or consistently failed are expected to provide less useful learning signal \citep{sukhbaatar2017intrinsic,florensa2018automatic}. 

Other methods seek to estimate the learning progress (LP) expected from training on a task \citep{baranes2013active}. Since future improvement cannot be observed in advance, most LP-based approaches estimate it from recent changes in competence and prioritize tasks or regions for which the estimated progress is greatest \citep{kanitscheider2021multi,forestier2022intrinsically}. Several variants use absolute learning progress (ALP), prioritizing tasks on which performance is deteriorating as well as improving \citep{matiisen2019teacher,colas2019curious}. These methods generally use recent changes in competence observed for a task as a proxy for the utility of further training on that task.

Recent work has explored combining such signals with LLMs to incorporate prior knowledge into curriculum construction. \citet{zhang2024omni} combine an LP-based curriculum with foundation-model judgments of task interestingness, while other approaches condition LLM task generation on task-specific estimates of the learner's current competence or learning progress \citep{pourcel2024autotelic,bolton2025sima}. These methods use LLM knowledge to identify tasks that are individually valuable and well suited to the learner's current state, but they do not model how training on one task may affect learning on others. Another line of work uses LLMs to decompose a complex target task into a sequence of subtasks that structure training toward it \citep{ryu2025curricullm,heravi2026leacl}. Such decompositions capture useful prerequisite structure, but are centered on a particular target and represent a more restricted class of relationship than general inter-task transfer.

A final line of work explicitly models inter-task transfer to guide curriculum construction. Some approaches approximate transfer using hand-designed proxies derived from available structure in the problem, including structural task descriptors and similarity or distance measures defined over a task representation \citep{svetlik2017automatic,silva2018object,tzannetos2024proximal}. Others estimate transfer from additional learner experience, including direct source-target transfer trials \citep{sinapov2015learning,jain2017faster},
cross-task evaluations during training \citep{huang2022curriculum}, or meta-learned models that predict the short-term learning progress induced on a target by training on different source tasks \citep{xu2023model}. Our approach instead leverages knowledge encoded in LLMs and their inference capabilities to obtain a prior over inter-task transfer, avoiding both the need to design problem-specific transfer proxies and the additional interaction required to estimate these relationships empirically.

\section{Problem formulation and analysis}\label{sec:Analysis and Motivation}

\subsection{The curriculum problem}
Automatic curriculum learning methods are concerned with the automatic selection of training data or situations so as to optimize learning. These selections can occur at different levels of granularity. Here, we focus on curricula that influence the tasks on which the agent collects new experience \citep{portelas2020automatic}. We represent each task \(m\) as a partially observable Markov decision process (POMDP), with tasks potentially differing in any of its components. Automatic curriculum learning can itself be viewed as a sequential decision-making problem \citep{narvekar2017autonomous,matiisen2019teacher}. The consequences of a curriculum decision depend on all information relevant to how the learner will respond to subsequent training. This information includes the learning problem specification (e.g., the tasks and objective of interest), the learning setup (e.g., the learning algorithm and its hyperparameters), and the current state of the learning process (e.g., learner and optimizer state). In principle, an ideal curriculum decision maker could condition its decisions on all such information. In practice, however, most of this information may be unavailable or difficult to exploit directly. Curriculum methods therefore generally operate on more tractable subsets or summaries, such as estimates of task performance or other statistics derived from the training history. 

For the analysis that follows, we formalize the curriculum process as a higher-level MDP. To facilitate the discussion and its interpretation, we adopt a simplified representation in which the state at curriculum decision $i$ corresponds to the learning agent's policy $\pi_i$, with initial state $\pi_0 \sim \rho_c$. The transition model $p_c(\pi_{i+1} \vert \pi_i, a_i)$ gives the probability that training policy $\pi_i$ on the task distribution induced by curriculum action $a_i$ results in policy $\pi_{i+1}$. We use $\pi_c$ to denote the curriculum policy that acts in this higher-level MDP, leaving any dependence on the decision index implicit. We consider a class of curriculum objectives that evaluate the learner after $N$ curriculum decisions:
\begin{equation}
J(\pi_c)
=
\mathbb{E}_{\rho_c,p_c,\pi_c}
\left[
\sum_m w(m) P(m;\pi_N)
\right].
\label{eq:curriculum-objective}
\end{equation}
Here, $P(m;\pi)$ is a performance measure of $\pi$ on task $m$ and the weight $w(m)\geq 0$ represents the importance assigned to task $m$. The expectation is taken over trajectories of the curriculum MDP induced by $\rho_c$, $p_c$, and $\pi_c$. Since $\rho_c$ is independent of the curriculum policy, maximizing $J\left(\pi_c\right)$ is equivalent to maximizing the improvement in performance over training. We define  $\operatorname{LP}\left(m, \pi_i, n\right):=P\left(m;\pi_{i+n}\right)-P\left(m;\pi_i\right)$ as the learning progress on task $m$ over $n$ curriculum decisions. We denote the equivalent objective by
\[
J_{\mathrm{LP}}(\pi_c) := \mathbb{E}_{\rho_c,p_c,\pi_c}\left[\sum_m w(m)\operatorname{LP}(m,\pi_0,N)\right] = \sum_{i=0}^{N-1}\sum_m w(m)\mathbb{E}_{\rho_c,p_c,\pi_c}\left[\operatorname{LP}(m,\pi_i,1)\right]
\]
The final expression makes explicit how each task's expected one-step learning progress contributes to the curriculum objective. We next analyze how task-selection criteria relate to this objective. We do so through the lens of LP, but the underlying considerations are more general.

\subsection{From local learning progress to optimal curriculum decisions}
Local LP-guided curriculum methods generally estimate the recent rate of change in performance on each task and use this information to prioritize subsequent training. Let $\widehat{\operatorname{LP}}_i(m)$ denote an estimate of the recent learning progress on task $m$ available before curriculum decision $i$. Methods differ in how these estimates are translated into task priority. We denote this transformation by $h$, such that task $m$ is prioritized according to $h(\widehat{\operatorname{LP}}_i(m))$, e.g., $h(x)=\max (x, 0)$ for positive-LP. For the remainder of the analysis, we consider curriculum decisions that allocate the next unit of training to a single task. Given policy $\pi_i$ and a task $m$ selected for training, we write $\operatorname{LP}(m',\pi_i,1;m):=P(m';\pi_{i+1})-P(m';\pi_i)$, with $\pi_{i+1}\sim p_c(\cdot\mid\pi_i,m)$, for the one-step learning progress induced on task $m'$. We propose that local LP-based methods can be interpreted as using $h(\widehat{\operatorname{LP}}_i(m))$ as a proxy for the expected one-step progress on $m$ resulting from allocating the next unit of training to $m$:
\begin{equation}
h\left(\widehat{\operatorname{LP}}_i(m)\right) \approx \mathbb{E}_{p_c\left(\pi_{i+1} \mid \pi_i, m\right)}\left[\operatorname{LP}\left(m, \pi_i, 1;m\right)\right].
\label{eq:lp-proxy}
\end{equation}
This interpretation allows us to analyze local LP-based selection in relation to the curriculum objective. If the proxy in Eq.~(\ref{eq:lp-proxy}) is sufficiently accurate to preserve the ordering across tasks, then the corresponding greedy rule selects the task whose performance is expected to improve the most if selected for training:
$
\argmax_m h\left(\widehat{\operatorname{LP}}_i(m)\right)=\argmax_m \mathbb{E}_{p_c\left(\pi_{i+1} \mid \pi_i, m\right)}\left[\operatorname{LP}\left(m, \pi_i, 1;m\right)\right].
$
This criterion is not generally optimal. To understand its limitations, we consider three sufficient conditions under which it is: (1) \emph{Equal task importance:} all tasks have the same objective weight $w(m)$; (2) \emph{No immediate cross-task effects}: training on $m$ does not immediately affect performance on any other task in expectation; and (3) \emph{No effect on future learning}: training on \(m\) does not change the expected learning progress obtainable from subsequent training decisions.

\paragraph{Varying task importance.}
Suppose Conditions 2 and 3 hold, but tasks may have different objective weights. When these weights are known, this limitation can be addressed simply by weighting the expected local improvement:
$
m_i^*=\argmax_m w(m) \mathbb{E}_{p_c\left(\pi_{i+1} \mid \pi_i, m\right)}\left[\operatorname{LP}\left(m, \pi_i, 1;m\right)\right]
$.

\paragraph{Immediate cross-task effects.}
A central source of curriculum learning's value is its ability to exploit cases where training on one task helps improve performance on others. Condition 2 excludes one important way in which such cross-task effects can arise, requiring 
$
\mathbb{E}_{p_c\left(\pi_{i+1} \mid \pi_i, m\right)}\left[\operatorname{LP}\left(m^{\prime}, \pi_i, 1;m\right)\right]=0
$
for all $m^{\prime} \neq m$. This is a particularly strong assumption in many deep learning settings, where a shared parameterized agent is trained across tasks and an update induced by experience on one task can consequently alter its behavior on the others. If Condition 2 is removed while Conditions 1 and 3 are retained, the optimal decision must account for the expected immediate effects on performance across all tasks:
\[
m_i^*=\argmax_m \Bigl[
\mathbb{E}_{p_c(\pi_{i+1}\mid\pi_i,m)}
[\operatorname{LP}(m,\pi_i,1;m)]
+\sum_{m'\neq m}
\mathbb{E}_{p_c(\pi_{i+1}\mid\pi_i,m)}
[\operatorname{LP}(m',\pi_i,1;m)]
\Bigr].
\]

\paragraph{Effects on future learning.}
The value of a curriculum decision can generally extend beyond the progress it produces immediately. Training on one task alters the learner's state, which can impact the effectiveness of future training. Condition 3 ignores such effects and the sequential nature of the curriculum problem. Under Conditions 1 and 2, the local criterion can be understood as greedily maximizing $J_{\mathrm{LP}}$ one iteration at a time. The optimal decision instead maximizes the expected progress accumulated over the full remaining horizon:
\[
m_i^*=\arg\max_m \Bigl[
\mathbb{E}_{p_c}\bigl[\operatorname{LP}(m,\pi_i,1;m)\bigr]
+
\mathbb{E}_{p_c,\pi_c^*}\biggl[
\sum_{j=1}^{N-i-1}\operatorname{LP}(m_{i+j},\pi_{i+j},1;m_{i+j})
\,\bigm|\,\pi_i,m_i=m
\biggr]
\Bigr].
\]

Dropping all three conditions jointly yields the following general expression for the optimal curriculum action at iteration \(i\):
\begin{equation}
\begin{aligned}
m_i^*
=\arg\max_m \Bigg[{}&
w(m)\,\mathbb{E}_{p_c}\bigl[\operatorname{LP}(m,\pi_i,1;m)\bigr]
+\mathbb{E}_{p_c}\biggl[\sum_{m'\neq m} w(m')\,\operatorname{LP}(m',\pi_i,1;m)\biggr]
\\
&+
\mathbb{E}_{p_c,\pi_c^*}\biggl[
\sum_{j=1}^{N-i-1}\sum_{m'} w(m')\,\operatorname{LP}(m',\pi_{i+j},1;m_{i+j})
\Bigm| \pi_i,m_i=m
\biggr]
\Bigg].
\end{aligned}
\label{eq:full-optimal-action}
\end{equation}

\paragraph{Historical LP as a proxy for current progress.}

The preceding analysis assumed that the approximation in Eq.~(\ref{eq:lp-proxy}) is sufficiently accurate. Estimates of past LP, however, inherit statistical noise from performance measurements and from the finite and potentially stale observations available for estimating recent changes in performance. Moreover, even an accurate estimate of recent past LP summarizes progress observed at previous learner states and therefore need not reflect the progress that training on $m$ would produce at the current policy $\pi_i$. In particular, for tasks on which no progress has yet been observed, past LP provides no basis for prioritization, even when their current expected learning progress differs.

In practice, LP-based curriculum methods use several mechanisms to address these limitations. The transformation $h$ applied to past LP can adjust how observed progress is translated into task priority. Absolute-LP criteria, for example, account for the discrepancy between past and future progress by treating recent deterioration as evidence that renewed training may be useful \citep{matiisen2019teacher,colas2019curious}. Methods also improve the LP estimate itself, for example, by refreshing performance estimates across the task set on every iteration at additional evaluation cost, or by aggregating performance observations over recent windows using moving averages or regression slopes \citep{matiisen2019teacher}. Finally, rather than selecting tasks greedily, methods commonly incorporate exploration through bandit-style selection, probabilistic sampling, or mixtures of LP-based and uniform sampling \citep{matiisen2019teacher, portelas2020teacher}.

\section{Method}\label{sec:Method}

Section~\ref{sec:Analysis and Motivation} highlights that the quality of a curriculum decision depends on consequences that are generally difficult to estimate. We seek to account for their influence without relying on additional learner experience for their estimation (see Section~\ref{sec:Related Work}). Commonly used curriculum signals, such as past performance estimates, are limited and may be insufficient for this purpose. We propose incorporating LLMs into curriculum decision making both as a flexible mechanism for reasoning about hard-to-estimate consequences affecting the value of curriculum actions and as a means of exploiting richer sources of information that may help infer their influence.

The proposal is intentionally general. LLMs can draw both on knowledge encoded in the model and on any supplied information about the curriculum problem or training history that admits a textual representation. They can support curriculum decision making in different ways and be used in both selection- and generation-based curricula. In this work, we develop a specific instantiation for task selection over a fixed and manageable task set \(\mathcal T\).
We further restrict the modeled downstream consequences of training to beneficial effects, omitting negative cross-task effects, and query the LLM only before training. Adaptation as learning proceeds relies on online task-performance estimates \(\widehat P_i(m)\in[0,1]\) and learning-progress estimates \(\widehat{\operatorname{LP}}_i(m)\) (Appendix~\ref{app:performance-lp-estimation}). These choices provide a controlled setting in which to study the proposed use of LLMs to enhance curriculum decisions, while limiting both the range of modeled effects and the computational cost of online decision making. In our experiments, we supply the LLM with textual descriptions of the tasks, the learner and training setup, and the intended use of its judgments in curriculum selection.

\subsection{Estimating downstream effects}
\label{sec:downstream-effects}

Equation~(\ref{eq:full-optimal-action}) shows that, beyond the direct contribution $w(m)\,\mathbb{E}_{p_c(\pi_{i+1}\mid\pi_i,m)}
\!\left[\operatorname{LP}(m,\pi_i,1;m)\right],$ the value of training on $m$ depends on immediate effects on other tasks and effects on future learning. Rather than attempting to estimate all these downstream terms explicitly, we construct a tractable approximation to their aggregate positive contribution. For downstream effects to arise, training on $m$ must impact the learner state. This motivates us to approximate their aggregate contribution through two factors: the extent to which next-step training on \(m\) is expected to induce relevant change, and the downstream value that such change has the potential to unlock. We consider $\mathbb{E}_{p_c(\pi_{i+1}\mid\pi_i,m)}
\!\left[\operatorname{LP}(m,\pi_i,1;m)\right]$ as a proxy for the former, and represent the latter through the \textbf{downstream potential} $D_i(m)$.

To estimate $D_i(m)$, we use an LLM before training to construct a directed \textbf{learning-connectivity graph} $\mathcal{G}_{\mathrm{LC}}$ (Appendix~\ref{app:graph-construction}). An edge $s\rightarrow t$, with LLM-derived weight $C(s,t)\in[0,1]$, represents evidence that useful learning on $s$ would either improve performance on $t$ directly or make subsequent training on $t$ more productive. We compute $D_i(m)$ by aggregating the contributions of tasks that useful learning on \(m\) is predicted to benefit. Each task \(t\) contributes according to its importance and the strength of the strongest path from \(m\). Let $\mathcal{P}_{i, L}(m, t)$ denote the paths from $m$ to $t$ of length at most $L$ in the adapted graph defined below. We define:
\[
D_i(m)
=
\sum_{
      \mathcal{P}_{i,L}(m,t) \neq \emptyset,\, t \neq m}
w(t)
\max_{p \in \mathcal{P}_{i,L}(m,t)}
\biggl[
\kappa^{|p|}
\prod_{(s,u) \in p} C_i(s,u)
\biggr],
\]
where $\kappa \in(0,1]$ downweights longer paths. Here, we adapt the graph to the current learner state by setting $C_i(s, t)=C(s, t) \mathbf{1}\left[\widehat{P}_i(t)<\tau_{\mathrm{solved}}\right]$, so that \(D_i(m)\) captures only potential downstream benefits associated with tasks that remain unsolved.

Finally, using \(h(\widehat{\operatorname{LP}}_i(m))\) as a nonnegative proxy for the unavailable expected next-step progress, following Eq.~(\ref{eq:lp-proxy}), yields the LP-based prioritization score
\[
S_i^{\mathrm{LP}}(m)=h\left(\widehat{\operatorname{LP}}_i(m)\right)\left[\widetilde{w}_i^{\mathrm{LP}}(m)+\alpha \widetilde{D}_i^{\mathrm{LP}}(m)\right].
\]
The first component recovers importance-weighted local LP prioritization, while the second additionally values progress according to the task's estimated downstream potential. Here, \(\widetilde{w}_i^{\mathrm{LP}}(m)\) and \(\widetilde{D}_i^{\mathrm{LP}}(m)\) are normalized versions of \(w(m)\) and \(D_i(m)\) over the current support of \(h(\widehat{\operatorname{LP}}_i(m))\). The normalization places the two quantities on comparable scales so that their relative influence can be controlled explicitly through \(\alpha\geq0\) (Appendix~\ref{app:score-normalization}).

\subsection{Estimating feasibility of progress}
\label{sec:feasibility-evidence}

Historical LP provides no evidence for distinguishing among tasks on which no success has yet been observed. We introduce $f_i(m)\in[0,1]$ to complement historical LP in estimating the expected progress from training on a low-performance task. Specifically, $f_i(m)$ estimates the current feasibility of making measurable progress by training directly on $m$.

To compute $f_i(m)$, we use an LLM before training to construct a directed \textbf{feasibility-evidence graph} $\mathcal{G}_{\mathrm{FE}}$ (Appendix~\ref{app:graph-construction}). For each target $m$, the LLM considers tasks $s$ on which it would ordinarily expect the learner to exhibit competence before direct training on $m$ begins producing measurable improvement. From these, it selects a small number of the most informative tasks, such that an edge $s\rightarrow m$, with nonnegative LLM-derived weight $E(s,m)$, represents competence on $s$ as evidence that $m$ is ready for productive training. The LLM may also judge \(m\) feasible from scratch, which we represent by an auxiliary source \(\bot\). At curriculum decision $i$, we combine these evidence weights with the learner's current performance estimates so that $f_i(m)$ increases as the learner becomes competent on tasks identified as evidence:
\[
f_i(m)
=
E(\bot,m)
+
\sum_{s\in\mathcal T\setminus\{m\}}
E(s,m)\widehat P_i(s).
\]
The incoming evidence weights are normalized to sum to one, ensuring \(f_i(m)\in[0,1]\).

Following the construction of \(S_i^{\mathrm{LP}}\) in Section~\ref{sec:downstream-effects}, we define
\[
S_i^f(m)
=
\mathbf{1}\!\left[\widehat P_i(m)\leq\tau_f\right]
f_i(m)
\left[
\widetilde w_i^f(m)
+
\alpha\,\widetilde D_i^f(m)
\right].
\]
Here, $f_i(m)$ serves in place of $h(\widehat{\operatorname{LP}}_i(m))$ as a proxy for $\mathbb{E}_{p_c(\pi_{i+1}\mid\pi_i,m)}
\!\left[\operatorname{LP}(m,\pi_i,1;m)\right]$, while the bracket values that progress according to its direct importance and estimated downstream potential. The performance threshold restricts this contribution to low-performance tasks. \(\widetilde{w}_i^f(m)\) and \(\widetilde{D}_i^f(m)\) are normalized versions of \(w(m)\) and \(D_i(m)\) over the current support of the thresholded feasibility signal (Appendix~\ref{app:score-normalization}).

\subsection{Task-selection distribution}
\label{sec:curriculum-distribution}

The scores introduced in Sections~\ref{sec:downstream-effects} and~\ref{sec:feasibility-evidence} provide complementary criteria for task prioritization. Both rely on imperfect estimates of the consequences of training on each task. As discussed in Section~\ref{sec:Analysis and Motivation}, this motivates probabilistic rather than fully greedy selection. A sampling distribution also naturally extends the selection mechanism to settings where multiple tasks are selected at each decision. We convert each score separately into a sampling distribution,\footnote{If the scores of a component sum to zero, its sampling distribution is defined to be uniform.} $q_i^{\mathrm{LP}}(m)
=
S_i^{\mathrm{LP}}(m) / \sum_{m'} S_i^{\mathrm{LP}}(m')$ and
$q_i^f(m)
=
S_i^f(m) / \sum_{m'} S_i^f(m')$. We also retain a uniform distribution $q^{\mathrm{U}}$ over the task set for score-independent exploration. After an initial period of uniform sampling, the curriculum task-selection distribution is
\[
q_i
=
\eta_{\mathrm{U}} q^{\mathrm{U}}
+
\eta_f q_i^f
+
\left(1-\eta_{\mathrm{U}}-\eta_f\right) q_i^{\mathrm{LP}},
\]
where $\eta_{\mathrm{U}},\eta_f\geq0$ and $\eta_{\mathrm{U}}+\eta_f\leq1$. The coefficients therefore directly control the proportion of training allocated through each component.

\section{Experiments}\label{sec:Experiments}
Our experiments evaluate whether incorporating LLM-informed downstream potential and feasibility guidance improves learning. We consider two settings: learning a designated target with access to auxiliary training tasks, and learning across an equally weighted task set.

\subsection{Benchmark and tasks}
We conduct experiments in Craftax \citep{matthews2024craftax}, a partially observable, procedurally generated environment designed for open-ended learning. In Craftax, the agent can traverse distinct biomes, gather and combine resources, craft and upgrade equipment, manage survival-related state, and interact with a diverse set of creatures. In our experiments, tasks differ only in their goal specification. Each task replaces the native Craftax rewards with a reward of one upon satisfying its goal and zero otherwise, with success terminating the episode. For this benchmark, \(P(m;\pi)\) is the probability that policy \(\pi\) achieves goal \(m\) within an episode. We introduce a custom set of 256 goals, each paired with a corresponding textual description. The set covers a broad space of objectives, with goals ranging from immediately accessible to those requiring long sequences of intermediate accomplishments. The complete goal set is provided in Appendix~\ref{app:benchmark-goals}. We use this task set under two kinds of curriculum objectives. In the first, one task is designated as the target, while the remaining tasks are available for auxiliary training. We consider four target tasks: reaching the Dungeon, defeating a skeleton, having at least 64 wood, and placing a furnace. We refer to them as Dungeon, Skeleton, Wood, and Furnace, respectively. In the second, all tasks receive equal weight, and performance is evaluated across the complete set.

\subsection{Learner and compared methods}
\label{sec:learner-compared-methods}

The compared curriculum methods differ only in how they allocate experience across tasks. In all experiments, the trained agent is a goal-conditioned policy optimized with proximal policy optimization \citep{schulman2017proximal}, consisting of a Gated Transformer-XL backbone \citep{dai2019transformer,parisotto2020stabilizing} with separate MLP heads for the actor and critic. The policy processes Craftax observations together with a fixed embedding of the textual goal produced by a frozen Qwen3-Embedding-4B model \citep{zhang2025qwen3}. Unless stated otherwise, the policy's action space consists of Craftax's primitive actions. For learning across the task set, we also evaluate a variant that can invoke its learned behavior for other goals as temporally bounded subroutines. Further details on the learner and training setup are provided in Appendices~\ref{app:learner-architecture}--\ref{app:policy-subroutine-calls}.

For the main method, we use \(h(x)=|x|^\rho\), with \(\rho>0\), and denote the resulting curriculum by \textbf{ALP+D+F}, where \textbf{D} denotes the inclusion of downstream potential when valuing progress on a goal and \textbf{F} the inclusion of the feasibility-guided distribution. The exponent \(\rho\) controls the sensitivity of the sampling distribution to differences in estimated absolute learning progress. We compare against variants that remove each of the two proposed components (\textbf{ALP+D} and \textbf{ALP+F}), and against \textbf{ALP}, an LLM-free baseline that adapts conventional ALP prioritization to the weighted objective and removes both components. We additionally compare against uniform goal sampling (\textbf{UNIF}) and an intermediate-difficulty curriculum (\textbf{I-DIFF}), which prioritizes goals for which current performance lies within a fixed competence range. In the single-target experiments, we also include \textbf{TARGET}, which allocates the entire training budget to the target goal, and \textbf{PERM}, an ablation of ALP+D+F that randomly permutes the goal labels of the learning-connectivity graph while leaving the feasibility-evidence graph unchanged. Full definitions are provided in Appendix~\ref{app:baselines-ablations}.

All adaptive curricula allocate 80\% of their sampling mass according to their primary prioritization score. For ALP, ALP+D, and I-DIFF, the remaining 20\% is assigned to uniform sampling; for ALP+F, ALP+D+F, and PERM, it is split equally between uniform and feasibility-guided sampling. Because feasibility-guided sampling can prioritize low-performance tasks before measurable progress is observed, we assign it part of the uniform exploration budget while keeping the LP-based allocation fixed. At each curriculum decision, we sample 64 goals with replacement from the resulting distribution and assign each to eight parallel environments, for a total of 512 environments. In the single-target experiments, for every method other than \textbf{TARGET}, four of the 64 goal selections are fixed to the target, with the remaining 60 sampled from the corresponding curriculum distribution. This guarantees all curriculum strategies a common minimum amount of direct training on the target. Further evaluation details are provided in Appendix~\ref{app:evaluation-procedure}.

\subsection{Results}

\subsubsection{Learning target tasks}
\label{sec:results-target-tasks}

\begin{figure}[htbp]
    \centering
    \includegraphics[width=\linewidth,trim=0bp 3bp 0bp 1bp,clip]{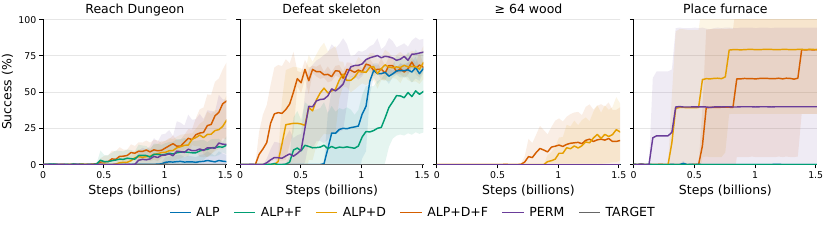}
    \caption{Evaluation success on selected target tasks during training, with all objective weight assigned to the corresponding target. Lines show means across five seeds; shaded regions indicate one standard deviation.}
    \label{fig:target-learning-curves}
\end{figure}

With all objective weight assigned to a single target, ALP+D+F substantially improves learning over ALP (Figure~\ref{fig:target-learning-curves}). After approximately 1.5 billion environment steps, it reaches mean success rates of 44\%, 17\%, and 79\% on Dungeon, Wood, and Furnace respectively, compared with 2\%, 0\%, and 0\% for ALP. On Skeleton, the advantage is primarily in learning speed: ALP+D+F reaches 50\% evaluation success earlier in every paired seed, while both methods finish with similar performance. TARGET has zero success at the final evaluation on all four targets. The onset of learning and final performance vary substantially across seeds.

The ablations indicate a stronger and more consistent benefit from downstream potential than from feasibility guidance. ALP+D improves over ALP on all four targets, and methods without D achieve no success by the end of training on Wood and Furnace. PERM retains some of the benefits of adding D, suggesting some tolerance to inaccuracies in the learning-connectivity graph. However, in its successful Dungeon and Furnace runs, we find that the permuted graphs still assign substantial downstream potential to prerequisite goals (approaching the ladder and obtaining a wooden pickaxe, respectively). Adding F to ALP and ALP+D leads to earlier improvements on Dungeon and Skeleton. For ALP+D, it also brings forward improvements on Wood but delays them on Furnace. These earlier gains do not consistently translate into higher final performance.

\begin{figure}[!b]
    \centering
    \includegraphics[width=0.96\linewidth,trim=0bp 3bp 0bp 1bp,clip]{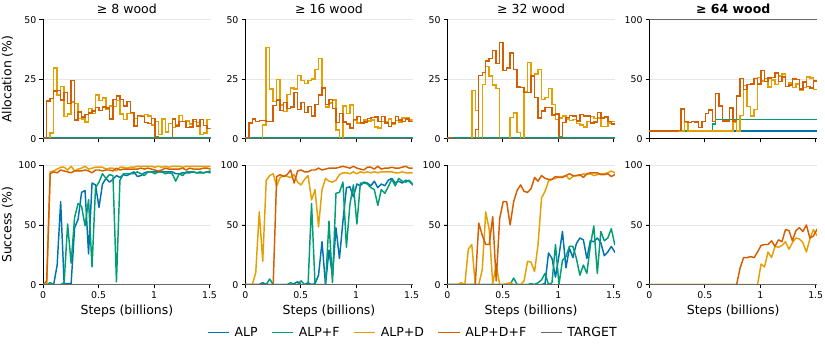}
    \caption{Illustrative curriculum dynamics with all objective weight assigned to having at least 64 wood. The top row shows the probability mass allocated to the target and selected related auxiliary goals, averaged over non-overlapping blocks of 15 curriculum decisions. The bottom row shows the policy's success rate when evaluated on each goal. For each method, we show the run with the highest final target success among the five seeds in Figure~\ref{fig:target-learning-curves}.}
    \label{fig:target-curriculum-wood}
\end{figure}

\paragraph{Curriculum dynamics}
Figure~\ref{fig:target-curriculum-wood} illustrates how the proposed components shape curriculum behavior on Wood using each method's highest-performing run. Appendix~\ref{app:curriculum-dynamics} contains illustrations for the other targets and all ALP+D+F seeds. Because only the target has nonzero objective weight, ALP and ALP+F can only adapt its sampling probability while all auxiliary goals remain equally likely. ALP distributes its adaptive mass uniformly in the absence of estimated target progress, and assigns it entirely to the target when progress is detected. Appendix~\ref{app:target-weight-smoothing} examines the effect of relaxing this restriction by assigning only half of the objective weight to the target. ALP+F can increase target probability earlier, once competence is acquired on the target's feasibility-evidence goals.  Here, learning on the 32-wood goal activates feasibility-guided sampling of the 64-wood target while target success remains zero.

Downstream potential changes the curriculum treatment of auxiliary training by explicitly valuing each task's predicted contribution to the target. This allows learning progress on related goals to guide task selection even before progress is observed on the target itself. In Figure~\ref{fig:target-curriculum-wood}, ALP+D and ALP+D+F devote substantial probability to lower wood thresholds while target success remains zero. The emphasis subsequently shifts toward direct target practice as performance on the lower thresholds becomes reliable and target success begins to increase. These trajectories illustrate how the task relationships supplied before training combine with online learning signals to prioritize potentially useful auxiliary goals and adapt their share of training as learning opportunities develop.

\subsubsection{Learning across the task set}
\label{sec:results-task-set}

\begin{figure}[htbp]
    \centering
    \includegraphics[width=\linewidth,trim=0bp 3bp 0bp 1bp,clip]{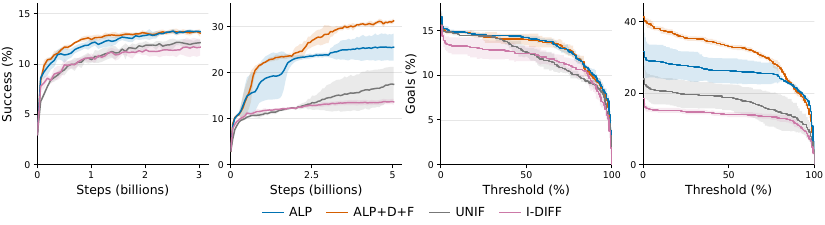}
    \caption{Evaluation under uniform task weights. The two left panels show mean success rate across all 256 tasks during training; the two right panels show the percentage of tasks whose success rate strictly exceeds each threshold after training. Within each pair, the policy restricted to primitive actions is shown first and the policy with subroutine calls second. Lines show means across five seeds; shaded regions indicate one standard deviation.}
    \label{fig:experiment-1-results}
\end{figure}

Under uniform task weights, ALP and ALP+D+F outperform both uniform sampling and intermediate-difficulty prioritization for both learner variants (Figure~\ref{fig:experiment-1-results}), supporting the use of learning progress to allocate training in this setting. For the learner restricted to primitive actions, adding downstream potential and feasibility guidance provides a modest advantage over ALP during the earlier stages of learning, but the methods reach similar performance by the end of training. Their final task coverage also overlaps strongly: about 94\% of the goals achieved by each method are also achieved by the other in paired runs, averaged across seeds. In this setting, the added components primarily bring improvements forward in training without materially extending final task coverage.

We additionally evaluate the curricula with a learner that can explicitly reuse behavior learned on other goals through temporally bounded subroutine calls. With this additional mechanism for inter-task transfer, the advantage of ALP+D+F over ALP is more pronounced and persists through the end of training. At the final evaluation, ALP+D+F reaches a mean success rate of 31.3\% across the task set, compared with 25.5\% for ALP, and achieves nonzero success on 41.3\% of goals, compared with 31.6\%. The results suggest that the value of accounting for the additional considerations captured by the LLM-informed components can depend on how learning transfers across tasks.

\section{Conclusion}\label{sec:Conclusion}
We analyzed automatic curriculum learning as a sequential decision making problem and examined several challenges involved in determining the value of curriculum actions. Motivated by this analysis, we investigated the use of LLMs to draw on richer information about the learning problem and better anticipate the consequences of curriculum decisions. We developed a concrete instantiation that augments learning-progress-based prioritization with LLM-informed estimates of the potential downstream benefits of training on a task and of whether direct training on a low-performance task is likely to produce progress. Our experiments in Craftax show that these components can improve learning and produce curriculum behavior consistent with their intended roles. The gains are most pronounced when optimizing for a single target and when learning across the full task set with a learner that can reuse goal-directed behaviors through subroutine calls. Our empirical evaluation is limited to a fixed set of goals, while the proposed instantiation uses offline LLM judgments and models only beneficial downstream effects. Future work could evaluate the approach on tasks that differ in aspects other than their goal specification, extend it to account for negative cross-task effects, and explore alternative ways of using LLMs to address the challenges identified by our analysis.

\bibliography{iclr2027_conference}
\bibliographystyle{iclr2027_conference}

\appendix
\etocdepthtag.toc{appendix}
\newpage

\begingroup
\renewcommand{\contentsname}{Appendix Contents}
\etocsettagdepth{main}{none}
\etocsettagdepth{appendix}{subsection}
\etocsetnexttocdepth{subsection}
\tableofcontents
\endgroup

\newpage

\section{Method details}
\label{app:detailed-method}

For the main method, we use a learning-progress exponent \(\rho=\tfrac12\), downstream coefficient \(\alpha=1\), maximum path length \(L=2\), path discount \(\kappa=0.3\), solved-task threshold \(\tau_{\mathrm{solved}}=0.9\), and feasibility threshold \(\tau_f=0.05\). Ablations retain these settings for the components they preserve. The downstream-potential and feasibility hyperparameters were fixed without tuning; tuning of shared hyperparameters is described in Appendix~\ref{app:hyperparameter-selection}, and sensitivity to \(L\) and \(\alpha\) is examined in Appendix~\ref{app:sensitivity-analysis}. Performance and learning-progress estimation settings are specified in Appendix~\ref{app:performance-lp-estimation}, while task-allocation settings are given in Section~\ref{sec:learner-compared-methods} and Appendix~\ref{app:baselines-ablations}.

\subsection{Pseudocode}
\label{app:algorithm-overview}

\begin{algorithm2e}[h]
\renewcommand{\NlSty}[1]{\textnormal{\textbf{\tiny #1}}}
    \caption{Curriculum procedure}
    \label{alg:online-curriculum}
    \KwIn{Task set $\mathcal{T}$; importance weights $w$; fixed graphs $\mathcal{G}_{\mathrm{LC}}$ and $\mathcal{G}_{\mathrm{FE}}$; initial learner $\pi_0$}
    Initialize the performance and learning-progress estimators\;
    \For{each curriculum decision $i$}{
        Update $\widehat P_i$ and $\widehat{\operatorname{LP}}_i$ using observations from the previous curriculum iteration, if available\;
        Adapt $\mathcal{G}_{\mathrm{LC}}$ to $\widehat P_i$ and compute the downstream potential $D_i$\;
        Compute the feasibility estimate $f_i$ from $\mathcal{G}_{\mathrm{FE}}$ and $\widehat P_i$\;
        Compute the scores $S_i^{\mathrm{LP}}$ and $S_i^f$\;
        Construct the task-selection distribution $q_i$, using uniform sampling during warm-up\;
        Sample tasks from $q_i$ and assign them to the training environments\;
        Train the learner on the assigned tasks and record the resulting performance observations\;
    }
\end{algorithm2e}

\subsection{Online performance and learning-progress estimation}
\label{app:performance-lp-estimation}

At each curriculum decision, the performance and learning-progress estimates are updated using training outcomes from the preceding curriculum iteration. After iteration \(i\), let \(\bar P_i(m)\) denote the empirical success rate across completed episodes in environments assigned task \(m\). For each environment, we include only its \(K=5\) most recent completed episodes.

Performance estimates are initialized as \(\widehat P_0(m)=0\). The first observation of a task replaces this initial value, while subsequent observations update the estimate using an exponential moving average:

\[
\widehat P_{i+1}(m)
=
\begin{cases}
\bar P_i(m),
& \text{if \(m\) is observed for the first time},\\[2mm]
(1-\beta)\widehat P_i(m)+\beta\bar P_i(m),
& \text{if \(m\) is observed again during iteration \(i\)},\\[2mm]
\widehat P_i(m),
& \text{if \(m\) is not observed during iteration \(i\)},
\end{cases}
\]

where \(\beta=0.3\).

Learning progress is estimated from the unsmoothed observations \(\bar P_i(m)\), rather than from \(\widehat P_i(m)\). For each task, we retain at most \(N_{\mathrm{obs}}=7\) observations, each paired with the curriculum iteration in which it was collected. When a new observation is available, \(\widehat{\operatorname{LP}}_{i+1}(m)\) is the slope of an ordinary least-squares line fitted to the current observation and the retained observations collected within the preceding \(H_{\mathrm{LP}}=12\) curriculum iterations. Thus, \(N_{\mathrm{obs}}\) limits the number of observations in the fit, whereas \(H_{\mathrm{LP}}\) limits the interval over which they may have been collected.

Each task history is initialized with the pseudo-observation \((-1,0)\). If no previous observation lies within the lookback interval, the most recent earlier observation is included and treated as occurring at the beginning of the interval. If no new observation is available for a task, its previous learning-progress estimate is retained.

\subsection{Score normalization}
\label{app:score-normalization}

Because the LP- and feasibility-based scores have different active supports, we normalize task importance and downstream potential separately for each component. Define

\[
\mathcal A_i^{\mathrm{LP}}
=
\left\{
m\in\mathcal T:
h\!\left(\widehat{\operatorname{LP}}_i(m)\right)>0
\right\},
\]

and

\[
\mathcal A_i^f
=
\left\{
m\in\mathcal T:
\mathbf 1\!\left[\widehat P_i(m)\leq\tau_f\right]f_i(m)>0
\right\}.
\]

For each component \(c\in\{\mathrm{LP},f\}\) with \(\mathcal A_i^c\neq\emptyset\), let

\[
\bar w_i^c
=
\frac{1}{|\mathcal A_i^c|}
\sum_{m\in\mathcal A_i^c}w(m),
\qquad
\bar D_i^c
=
\frac{1}{|\mathcal A_i^c|}
\sum_{m\in\mathcal A_i^c}D_i(m).
\]

The normalized quantities used in the corresponding score are

\[
\widetilde w_i^c(m)
=
\frac{w(m)}{\bar w_i^c},
\qquad
\widetilde D_i^c(m)
=
\frac{D_i(m)}
{\max\!\left\{\bar D_i^c,\bar w_i^c\right\}}.
\]

When \(\bar D_i^c\geq\bar w_i^c>0\), each quantity is normalized by its own active-set mean, giving the direct and downstream terms the same mean scale before weighting by \(\alpha\). When \(\bar D_i^c<\bar w_i^c\), the downstream term is instead normalized by \(\bar w_i^c\), preventing a small estimated downstream signal from being amplified to the scale of the direct contribution solely through normalization.\footnote{If \(\mathcal A_i^c=\emptyset\), \(S_i^c\) is defined to be identically zero. If a denominator in the normalization above is zero, the corresponding normalized quantity is defined to be zero.}

\section{Offline LLM graph generation}
\label{app:graph-construction}
\newcommand{\graphpromptinput}[1]{%
  \VerbatimInput[
    fontencoding=T1,
    breaklines=true,
    breaksymbolleft={},
    breaksymbolright={},
    fontsize=\scriptsize,
    baselinestretch=0.92,
    frame=single,
    framerule=0.3pt,
    rulecolor=\color{black!25},
    fillcolor=\color{black!2},
    framesep=2mm
  ]{#1}%
}

\subsection{Construction procedure}
\label{app:graph-generation}

Both graphs are constructed offline over the 256 benchmark goals from repeated
LLM judgments about which directed edges should be included. Each query provides
the LLM with descriptions of the Craftax environment, goal suite, learner, and
training setting, together with graph-specific instructions and a structured
response schema. For each graph type, we generate \(R=5\) versions using
different goal orderings and aggregate their judgments into the edge weights
used during training. We use Gemini 3.7 Flash with thinking level set to high.
Temperature and top-\(p\) are left at their API defaults.

\paragraph{Learning-connectivity graph.}
Every goal is considered both as a potential source and as a potential target.
In each repetition, we independently shuffle two copies of the complete goal
set, one for each role. The source goals are divided into blocks of at most 64
and the target goals into blocks of at most 16, and we query every pairing of a
source block with a target block. Consequently, every ordered pair of distinct
goals is considered exactly once per repetition.

Let \(C^{(r)}(s,t)\in\{0,1\}\) indicate whether the LLM includes the edge
\(s\rightarrow t\) in repetition \(r\), with self-connections excluded. The
final connectivity weight is

\begin{equation}
C(s,t)
=
\frac{1}{R}\sum_{r=1}^{R} C^{(r)}(s,t).
\label{eq:lc-graph-aggregation}
\end{equation}

Thus, \(C(s,t)\) is the fraction of repetitions in which the LLM includes the
edge.

\paragraph{Feasibility-evidence graph.}
For each goal \(m\), we evaluate all other goals as potential evidence sources.
In each repetition, we shuffle the order of the potential evidence goals and
independently shuffle the target goals. The targets are divided into blocks of
at most eight, and each query evaluates one target block while providing the
complete goal list as evidence candidates; the evaluated goal itself cannot be
selected. Thus, every goal is evaluated exactly once per repetition with all
other goals available as potential evidence.

For each target \(m\), the LLM either judges direct training feasible from
initialization or selects between one and three other goals as evidence. Let
\(A^{(r)}(m)\) contain \(\bot\) in the former case and the selected evidence
goals otherwise. We convert each judgment into edge weights by assigning total
weight one across the members of \(A^{(r)}(m)\): each selected source receives
weight \(1/|A^{(r)}(m)|\), and all other sources receive zero. The final
feasibility-evidence weight averages these values across repetitions:

\begin{equation}
E(s,m)
=
\frac{1}{R}
\sum_{r=1}^{R}
\frac{\mathbf{1}\!\left[s\in A^{(r)}(m)\right]}
     {|A^{(r)}(m)|},
\qquad
s\in(\mathcal{T}\setminus\{m\})\cup\{\bot\}.
\label{eq:fe-graph-aggregation}
\end{equation}

Dividing the weight equally prevents a repetition from contributing more total
evidence merely because it selects more sources. It also ensures

\begin{equation}
E(\bot,m)+\sum_{s\in\mathcal T\setminus\{m\}}E(s,m)=1
\label{eq:fe-graph-normalization}
\end{equation}
for every target \(m\).

\subsection{Prompts}
\label{app:graph-prompts}

The natural-language prompt templates used to construct the two graphs are
reproduced below. Each query instantiates the relevant template with three
context blocks describing the environment, goal suite, and learner,
together with the query-specific goal lists determined by the batching procedure
in Appendix~\ref{app:graph-generation}. In our experiments, we use these templates
to generate one pair of graphs for each learner variant, changing only the
learner context. Each pair is then reused across runs with the corresponding
learner.

Angle-bracketed terms denote these
inserted fields. For learning-connectivity queries, \texttt{<SOURCE GOALS>} and
\texttt{<TARGET GOALS>} contain the corresponding source and target blocks. For
feasibility-evidence queries, \texttt{<EVIDENCE GOALS>} contains the complete
goal set and \texttt{<TARGET GOALS>} the current target block. Each query was
additionally supplied with a graph-specific JSON response schema used to
constrain the model output to the required structured format; the schema
definitions are omitted here.

\paragraph{Learning-connectivity prompt.}
The template used for each learning-connectivity query is:

\graphpromptinput{sections/prompts/lc_graph_prompt.txt}

\paragraph{Feasibility-evidence prompt.}
The template used for each feasibility-evidence query is:

\graphpromptinput{sections/prompts/feasibility_graph_prompt.txt}

\label{app:shared-prompt-components}

\paragraph{Shared context.}
The following environment, goal-suite, and learner descriptions are inserted
into the corresponding placeholders in both prompt templates. The environment
and goal-suite descriptions are shared across learner variants, while the
learner description depends on the variant.

\paragraph{Environment context.}
\graphpromptinput{sections/prompts/craftax_environment_context.txt}

\paragraph{Goal-suite context.}
\graphpromptinput{sections/prompts/goal_suite_context.txt}

\paragraph{Learner context.}\mbox{}\par
\smallskip
\noindent{\small\itshape Policy restricted to primitive actions.}\par
\graphpromptinput{sections/prompts/learner_context.txt}

\smallskip
\noindent{\small\itshape Policy with subroutine calls.}\par
\graphpromptinput{sections/prompts/learner_subroutine_context.txt}

\section{Experimental setup}
\label{app:experimental-setup}

\subsection{Benchmark goal set}
\label{app:benchmark-goals}

The benchmark pairs textual goal descriptions with success conditions evaluated directly from environment states or transitions. The complete set of 256 goals is presented in Table~\ref{tab:benchmark-goals}. For spatial objectives, ``See'' requires the target to be visible within the agent's local observation window, accounting for lighting, whereas ``Be adjacent'' requires it to occupy one of the four cardinally neighbouring tiles. Inventory thresholds specify the quantity that must be held at one time, rather than the total collected over an episode.

\begingroup
\small
\setlength{\LTcapwidth}{\textwidth}
\begin{longtable}{@{}>{\raggedright\arraybackslash}p{0.62\textwidth}>{\raggedright\arraybackslash}p{0.20\textwidth}>{\raggedleft\arraybackslash}p{\dimexpr0.18\textwidth-4\tabcolsep\relax}@{}}
\caption{Complete 256-goal benchmark. A row containing \(n\) represents one goal for each listed value; countable nouns take the singular form when \(n=1\). All other rows represent one goal. The cumulative count gives the number of goals represented up to and including each row. Goal specifications reproduce the canonical goal descriptions, except that scalar threshold values are replaced by \(n\).}
\label{tab:benchmark-goals}\\
\toprule
Goal specification & Values of \(n\) & Cumulative count \\
\midrule
\endfirsthead

\multicolumn{3}{l}{\small\textit{Table~\ref{tab:benchmark-goals} continued}}\\
\toprule
Goal specification & Values of \(n\) & Cumulative count \\
\midrule
\endhead

\midrule
\multicolumn{3}{r}{\small\textit{Continued on next page}}\\
\endfoot

\bottomrule
\endlastfoot

Have at least \(n\) wood. & \(1, 2, 4, 8, 16, 32, 64\) & 7 \\
Have at least \(n\) stone. & \(1, 2, 4, 8, 16, 32, 64\) & 14 \\
Have at least \(n\) coal. & \(1, 2, 4, 8, 12, 24, 48\) & 21 \\
Have at least \(n\) iron. & \(1, 2, 3, 6, 12, 24\) & 27 \\
Have at least \(n\) diamonds. & \(1, 2, 3, 6, 12\) & 32 \\
Have at least \(n\) sapphires. & \(1, 2, 4, 8\) & 36 \\
Have at least \(n\) rubies. & \(1, 2, 4, 8\) & 40 \\
Have at least \(n\) saplings. & \(1, 2, 4, 8\) & 44 \\
Have at least \(n\) torches. & \(4, 8, 16, 32, 64\) & 49 \\
Have at least \(n\) arrows. & \(2, 4, 8, 16, 32\) & 54 \\
Have at least \(n\) books. & \(1, 2\) & 56 \\
Obtain a wood pickaxe or better. & \textemdash & 57 \\
Obtain a stone pickaxe or better. & \textemdash & 58 \\
Obtain an iron pickaxe or better. & \textemdash & 59 \\
Obtain a diamond pickaxe. & \textemdash & 60 \\
Obtain a wood sword or better. & \textemdash & 61 \\
Obtain a stone sword or better. & \textemdash & 62 \\
Obtain an iron sword or better. & \textemdash & 63 \\
Obtain a diamond sword. & \textemdash & 64 \\
Obtain a bow. & \textemdash & 65 \\
Obtain one piece of iron armour or better. & \textemdash & 66 \\
Obtain one piece of diamond armour. & \textemdash & 67 \\
Obtain an iron chestplate or better. & \textemdash & 68 \\
Obtain a diamond chestplate. & \textemdash & 69 \\
Obtain iron leggings or better. & \textemdash & 70 \\
Obtain diamond leggings. & \textemdash & 71 \\
Obtain iron boots or better. & \textemdash & 72 \\
Obtain diamond boots. & \textemdash & 73 \\
Apply a fire enchantment to a sword. & \textemdash & 74 \\
Apply an ice enchantment to a sword. & \textemdash & 75 \\
Apply a fire enchantment to a bow. & \textemdash & 76 \\
Apply an ice enchantment to a bow. & \textemdash & 77 \\
Apply a fire enchantment to one piece of armour. & \textemdash & 78 \\
Apply an ice enchantment to one piece of armour. & \textemdash & 79 \\
Reach dexterity level \(n\). & \(2, 3, 4, 5\) & 83 \\
Reach strength level \(n\). & \(2, 3, 4, 5\) & 87 \\
Reach intelligence level \(n\). & \(2, 3, 4, 5\) & 91 \\
Reach floor 1, the Dungeon. & \textemdash & 92 \\
Reach floor 2, the Gnomish Mines. & \textemdash & 93 \\
Reach floor 3, the Sewers. & \textemdash & 94 \\
Reach floor 4, the Vault. & \textemdash & 95 \\
Reach floor 5, the Troll Mines. & \textemdash & 96 \\
Reach floor 6, the Fire Realm. & \textemdash & 97 \\
Reach floor 7, the Ice Realm. & \textemdash & 98 \\
Reach floor 8, the Graveyard. & \textemdash & 99 \\
Place a crafting table. & \textemdash & 100 \\
Place a furnace. & \textemdash & 101 \\
Plant a sapling. & \textemdash & 102 \\
See coal. & \textemdash & 103 \\
See iron. & \textemdash & 104 \\
See diamond. & \textemdash & 105 \\
See lava. & \textemdash & 106 \\
See a mossy wall. & \textemdash & 107 \\
See sapphire. & \textemdash & 108 \\
See ruby. & \textemdash & 109 \\
See a chest. & \textemdash & 110 \\
See a fountain. & \textemdash & 111 \\
See a fire enchantment table. & \textemdash & 112 \\
See an ice enchantment table. & \textemdash & 113 \\
See the downward ladder on floor 0, the Overworld. & \textemdash & 114 \\
See the downward ladder on floor 1, the Dungeon. & \textemdash & 115 \\
See the downward ladder on floor 2, the Gnomish Mines. & \textemdash & 116 \\
See the downward ladder on floor 3, the Sewers. & \textemdash & 117 \\
See the downward ladder on floor 4, the Vault. & \textemdash & 118 \\
See the downward ladder on floor 5, the Troll Mines. & \textemdash & 119 \\
See the downward ladder on floor 6, the Fire Realm. & \textemdash & 120 \\
See the downward ladder on floor 7, the Ice Realm. & \textemdash & 121 \\
See a cow. & \textemdash & 122 \\
See a zombie. & \textemdash & 123 \\
See a skeleton. & \textemdash & 124 \\
See a snail. & \textemdash & 125 \\
See an orc soldier. & \textemdash & 126 \\
See an orc mage. & \textemdash & 127 \\
See a bat. & \textemdash & 128 \\
See a gnome warrior. & \textemdash & 129 \\
See a gnome archer. & \textemdash & 130 \\
See a lizard. & \textemdash & 131 \\
See a kobold. & \textemdash & 132 \\
See a knight. & \textemdash & 133 \\
See an archer. & \textemdash & 134 \\
See a troll. & \textemdash & 135 \\
See a Deep Thing. & \textemdash & 136 \\
See a pigman. & \textemdash & 137 \\
See a fire elemental. & \textemdash & 138 \\
See a frost troll. & \textemdash & 139 \\
See an ice elemental. & \textemdash & 140 \\
See the Necromancer. & \textemdash & 141 \\
Be adjacent to water. & \textemdash & 142 \\
Be adjacent to stone. & \textemdash & 143 \\
Be adjacent to a path. & \textemdash & 144 \\
Be adjacent to sand. & \textemdash & 145 \\
Be adjacent to coal. & \textemdash & 146 \\
Be adjacent to iron. & \textemdash & 147 \\
Be adjacent to diamond. & \textemdash & 148 \\
Be adjacent to lava. & \textemdash & 149 \\
Be adjacent to a ripe plant. & \textemdash & 150 \\
Be adjacent to a torch. & \textemdash & 151 \\
Be adjacent to a wall. & \textemdash & 152 \\
Be adjacent to a mossy wall. & \textemdash & 153 \\
Be adjacent to a stalagmite. & \textemdash & 154 \\
Be adjacent to sapphire. & \textemdash & 155 \\
Be adjacent to ruby. & \textemdash & 156 \\
Be adjacent to a chest. & \textemdash & 157 \\
Be adjacent to a fountain. & \textemdash & 158 \\
Be adjacent to fire grass. & \textemdash & 159 \\
Be adjacent to a fire tree. & \textemdash & 160 \\
Be adjacent to a fire enchantment table. & \textemdash & 161 \\
Be adjacent to ice grass. & \textemdash & 162 \\
Be adjacent to an ice shrub. & \textemdash & 163 \\
Be adjacent to an ice enchantment table. & \textemdash & 164 \\
Be adjacent to a grave. & \textemdash & 165 \\
Be adjacent to the downward ladder on floor 0, the Overworld. & \textemdash & 166 \\
Be adjacent to the downward ladder on floor 1, the Dungeon. & \textemdash & 167 \\
Be adjacent to the downward ladder on floor 2, the Gnomish Mines. & \textemdash & 168 \\
Be adjacent to the downward ladder on floor 3, the Sewers. & \textemdash & 169 \\
Be adjacent to the downward ladder on floor 4, the Vault. & \textemdash & 170 \\
Be adjacent to the downward ladder on floor 5, the Troll Mines. & \textemdash & 171 \\
Be adjacent to the downward ladder on floor 6, the Fire Realm. & \textemdash & 172 \\
Be adjacent to the downward ladder on floor 7, the Ice Realm. & \textemdash & 173 \\
Be adjacent to a cow. & \textemdash & 174 \\
Be adjacent to a zombie. & \textemdash & 175 \\
Be adjacent to a skeleton. & \textemdash & 176 \\
Be adjacent to a snail. & \textemdash & 177 \\
Be adjacent to an orc soldier. & \textemdash & 178 \\
Be adjacent to an orc mage. & \textemdash & 179 \\
Be adjacent to a bat. & \textemdash & 180 \\
Be adjacent to a gnome warrior. & \textemdash & 181 \\
Be adjacent to a gnome archer. & \textemdash & 182 \\
Be adjacent to a lizard. & \textemdash & 183 \\
Be adjacent to a kobold. & \textemdash & 184 \\
Be adjacent to a knight. & \textemdash & 185 \\
Be adjacent to an archer. & \textemdash & 186 \\
Be adjacent to a troll. & \textemdash & 187 \\
Be adjacent to a Deep Thing. & \textemdash & 188 \\
Be adjacent to a pigman. & \textemdash & 189 \\
Be adjacent to a fire elemental. & \textemdash & 190 \\
Be adjacent to a frost troll. & \textemdash & 191 \\
Be adjacent to an ice elemental. & \textemdash & 192 \\
Be adjacent to the Necromancer. & \textemdash & 193 \\
Damage a cow. & \textemdash & 194 \\
Damage a zombie. & \textemdash & 195 \\
Damage a skeleton. & \textemdash & 196 \\
Damage a snail. & \textemdash & 197 \\
Damage an orc soldier. & \textemdash & 198 \\
Damage an orc mage. & \textemdash & 199 \\
Damage a bat. & \textemdash & 200 \\
Damage a gnome warrior. & \textemdash & 201 \\
Damage a gnome archer. & \textemdash & 202 \\
Damage a lizard. & \textemdash & 203 \\
Damage a kobold. & \textemdash & 204 \\
Damage a knight. & \textemdash & 205 \\
Damage an archer. & \textemdash & 206 \\
Damage a troll. & \textemdash & 207 \\
Damage a Deep Thing. & \textemdash & 208 \\
Damage a pigman. & \textemdash & 209 \\
Damage a fire elemental. & \textemdash & 210 \\
Damage a frost troll. & \textemdash & 211 \\
Damage an ice elemental. & \textemdash & 212 \\
Defeat a cow. & \textemdash & 213 \\
Defeat a zombie. & \textemdash & 214 \\
Defeat a skeleton. & \textemdash & 215 \\
Defeat a snail. & \textemdash & 216 \\
Defeat an orc soldier. & \textemdash & 217 \\
Defeat an orc mage. & \textemdash & 218 \\
Defeat a bat. & \textemdash & 219 \\
Defeat a gnome warrior. & \textemdash & 220 \\
Defeat a gnome archer. & \textemdash & 221 \\
Defeat a lizard. & \textemdash & 222 \\
Defeat a kobold. & \textemdash & 223 \\
Defeat a knight. & \textemdash & 224 \\
Defeat an archer. & \textemdash & 225 \\
Defeat a troll. & \textemdash & 226 \\
Defeat a Deep Thing. & \textemdash & 227 \\
Defeat a pigman. & \textemdash & 228 \\
Defeat a fire elemental. & \textemdash & 229 \\
Defeat a frost troll. & \textemdash & 230 \\
Defeat an ice elemental. & \textemdash & 231 \\
Kill at least \(n\) hostile monsters on floor 1, the Dungeon. & \(1, 2, 4, 8\) & 235 \\
Kill at least \(n\) hostile monsters on floor 2, the Gnomish Mines. & \(1, 2, 4, 8\) & 239 \\
Kill at least \(n\) hostile monsters on floor 3, the Sewers. & \(1, 2, 4, 8\) & 243 \\
Kill at least \(n\) hostile monsters on floor 4, the Vault. & \(1, 2, 4, 8\) & 247 \\
Kill at least \(n\) hostile monsters on floor 5, the Troll Mines. & \(2, 4, 8\) & 250 \\
Kill at least \(n\) hostile monsters on floor 6, the Fire Realm. & \(2, 8\) & 252 \\
Kill at least \(n\) hostile monsters on floor 7, the Ice Realm. & \(2, 8\) & 254 \\
Kill at least \(n\) hostile monsters on floor 8, the Graveyard. & \(2, 8\) & 256 \\
\midrule
\multicolumn{2}{r}{Total} & 256 \\
\end{longtable}
\endgroup

The success evaluator can sometimes fail to register a lethal projectile or spell attack on a cow, bat, or snail. For the corresponding damage and defeat goals, the agent may therefore need to repeat the interaction with another creature of the same type before success is registered.

\subsection{Learner architecture}
\label{app:learner-architecture}

The observation encoder processes the spatial and non-spatial components of the symbolic Craftax observation separately. For each spatial cell, learned embeddings represent blocks, items, creatures and projectiles, and visibility. These embeddings are concatenated and projected to 32 features, then processed by two \(3\times3\) convolutional layers with 32 channels each. The non-spatial features pass through a fully connected layer with 64 units. The flattened spatial features and encoded non-spatial features are concatenated and projected to a 512-dimensional observation embedding. GELU activations follow each fully connected and convolutional layer in the observation encoder.

Each goal description is encoded offline using a frozen Qwen3-Embedding-4B model, with embeddings truncated to 512 dimensions and normalized to unit length. These embeddings remain fixed throughout training. Both learners use a Gated Transformer-XL backbone consisting of two blocks with a hidden dimension of 256, four attention heads, and a memory length of 64 steps. Gate biases are initialized to 2. During training, backpropagation is truncated to subsequences of 64 steps, with preceding cached memory states treated as constants. The actor and critic use separate MLP heads, each with two hidden layers of 512 units and ReLU activations.

For the policy restricted to primitive actions, the observation and goal embeddings are concatenated and passed through a fully connected layer with 512 units and a GELU activation, followed by a linear projection to the Transformer hidden dimension. The same goal embedding is also concatenated with the Transformer output before each actor and critic head. The actor assigns probabilities to the 43 primitive actions. The complete actor--critic network contains approximately 6.17 million trainable parameters, excluding the frozen text encoder.

For the policy with subroutine calls, the observation embedding is projected directly to the Transformer hidden dimension, without goal conditioning. The actor receives the Transformer output concatenated with the embedding of the goal assigned to the episode or, during a subroutine, the currently invoked goal. The critic additionally receives the embedding of the goal assigned to the episode and the remaining delegation time, expressed as a fraction of the maximum delegation duration. The actor assigns probabilities to the 43 primitive actions and 256 call actions, one per benchmark goal. The complete actor--critic network contains approximately 6.04 million trainable parameters, excluding the frozen text encoder.

\subsection{Training protocol}
\label{app:training-protocol}

At the start of each curriculum iteration, all environments are reset. Each environment then runs for 4,096 steps with a fixed assigned goal. Episodes are limited to 4,096 steps. After every 256 steps per environment, we perform one epoch of PPO optimization over eight minibatches.

We use PPO with a clipping parameter of \(0.2\) and the same threshold for value-function clipping. Advantages are computed using generalized advantage estimation (GAE) with \(\gamma=0.999\) and \(\lambda=0.97\). For the policy restricted to primitive actions, advantages are normalized to zero mean and unit variance across the complete rollout batch. The value-loss coefficient is \(0.5\), and the entropy coefficient is \(0.005\). The changes to the actor loss, advantage normalization, and regularization for the policy with subroutine calls are described in Appendix~\ref{app:policy-subroutine-calls}.

We optimize the network using Adam with a constant learning rate of \(2\times10^{-4}\) and \(\epsilon=10^{-5}\). Gradients are clipped to a maximum global norm of \(1\).

\subsection{Policy with subroutine calls}
\label{app:policy-subroutine-calls}

The curriculum assigns each episode a goal \(g\in\mathcal T\). The agent receives a reward of one upon achieving \(g\) and zero otherwise, with success ending the episode. This policy variant augments the primitive Craftax action set \(\mathcal A_{\mathrm{prim}}\) with one call action per benchmark goal, giving \(\mathcal A_{\mathrm{combined}}=\mathcal A_{\mathrm{prim}}\cup\{\operatorname{call}(g'):g'\in\mathcal T\}\). The actor assigns probabilities over this combined action space, so invoking a goal is selected in the same way as a primitive action. These calls provide an explicit mechanism for reusing behavior learned while training on other goals.

Selecting a primitive action executes it directly in the environment. Selecting \(\operatorname{call}(g')\) executes one no-op environment step, after which the policy switches its conditioning to \(g'\). It returns to conditioning on \(g\) after 32 further environment steps. During these 32 steps, the policy can select either primitive actions or further call actions. Each call consumes a no-op step and replaces the goal used for the remaining steps, without extending the original deadline. Execution then returns directly to \(g\). A complete call block therefore occupies 33 environment steps, regardless of calls made within it. All steps count toward the episode and training budgets. The same network generates actions throughout. Environment state and Transformer memory persist across changes in goal conditioning. Achieving a goal other than \(g\) provides no reward and does not cause an early return. Episode termination ends any ongoing subroutine.

At the start of curriculum iteration \(i\), calls are restricted to goals in \(\mathcal E_i=\{g'\in\mathcal T:\widehat P_i(g')\geq 1/3\}\) to avoid spending time invoking goals on which the policy has demonstrated little competence. This set remains fixed throughout the iteration and its PPO updates. Primitive actions are always available, and the curriculum remains free to select any goal for training.

The PPO actor loss uses only decisions made outside subroutines, including actions that initiate calls, and actor advantages are normalized over these decisions. GAE and critic learning use all environment steps, with returns defined by the episode's assigned goal \(g\). Rewards obtained during a subroutine can therefore inform the update to the decision that invoked it, while actions taken within the subroutine receive no direct actor loss.

Assigning equal probability to every available action would make subroutine calls more frequent as more goals become eligible. It would also fail to account for the fact that a complete call occupies 33 environment steps, whereas a primitive action occupies one. We therefore define a reference distribution \(\mu_i\) that, when calls are available, assigns total probability \(33/34\) to primitive actions and \(1/34\) to eligible calls, distributed uniformly within each group. Assuming calls run to completion, this choice gives equal expected environment time to direct primitive actions and call blocks. When no goals are eligible, \(\mu_i\) is uniform over primitive actions.

The actor learns preferences relative to this reference distribution. Given its output logit \(\ell_t(a)\) for action \(a\), action probabilities are \(\pi_t(a)\propto\mu_i(a)\exp(\ell_t(a))\). Equal logits recover the reference distribution, while learned preferences can change the allocation between primitive actions and calls. The same construction is used both outside and during subroutines.

The ordinary entropy bonus favors uniform probabilities over available actions. To make regularization consistent with the chosen reference distribution, we replace this bonus with a term equivalent to penalizing \(D_{\mathrm{KL}}(\pi_t\Vert\mu_i)\). This penalty is averaged over decisions made outside subroutines and uses the same coefficient as the entropy bonus specified in Appendix~\ref{app:training-protocol}. When no goals are eligible, this recovers ordinary entropy regularization.

\subsection{Evaluation procedure}
\label{app:evaluation-procedure}

For the reported results, policies are evaluated every 15 curriculum decisions in Section~\ref{sec:results-target-tasks} and every 30 in Section~\ref{sec:results-task-set}. At each evaluation point, each goal's success rate is estimated from 128 separate evaluation episodes for each training seed, with actions sampled from the current policy. Each comparison uses five training seeds paired across methods. Aggregate performance curves show the mean across seeds, with shaded regions indicating one standard deviation.

For the policy with subroutine calls, the set of callable goals is determined from the latest training performance estimates and remains fixed throughout evaluation. Evaluation outcomes do not update these estimates.

Evaluating every benchmark goal as a separate single-target objective was computationally prohibitive. After the uniform-weight experiments for the policy restricted to primitive actions, we selected candidate targets spanning qualitatively different objectives and judged to be challenging but plausibly learnable within the available training budget. For each candidate, we ran TARGET, ALP, ALP+D, ALP+F, and ALP+D+F with a shared training seed and retained targets for which at least one method achieved nonzero success at the final evaluation. This yielded the Dungeon, Skeleton, Wood, and Furnace targets; ``See an orc soldier'' and ``Obtain a stone sword or better'' were excluded under this criterion. These initial runs were included in the reported five-seed comparisons.

\subsection{Hyperparameter selection}
\label{app:hyperparameter-selection}

Most hyperparameters were fixed without tuning. All hyperparameter tuning was conducted using the ALP baseline on the complete 256-goal benchmark, with uniform task weights and the policy restricted to primitive actions. The selected values were then shared across curriculum methods and learner variants wherever applicable. Selection proceeded through successive comparisons against a reference configuration, usually changing one setting at a time. Most candidates were evaluated on one development seed, with selected comparisons repeated on a second seed. Runs were generally allocated approximately two billion environment steps, and all development seeds were distinct from those used in the main experiments. Tuning used a smaller variant of the final network; the architecture used in the main experiments was evaluated separately. Table~\ref{tab:hyperparameter-selection} reports the candidate and selected values.

\begin{table}[htbp]
\centering
\caption{Numerical hyperparameters considered during tuning. Values were evaluated through successive comparisons rather than a joint grid search.}
\label{tab:hyperparameter-selection}
\small
\renewcommand{\arraystretch}{1.1}
\begin{tabular}{@{}>{\raggedright\arraybackslash}p{0.42\linewidth}>{\centering\arraybackslash}p{0.41\linewidth}>{\centering\arraybackslash}p{\dimexpr0.17\linewidth-4\tabcolsep\relax}@{}}
\toprule
Hyperparameter & Values evaluated & Selected \\
\midrule
Learning rate & \(3\times10^{-5},\,5\times10^{-5},\,10^{-4},\,2\times10^{-4}\) & \(2\times10^{-4}\) \\
Entropy coefficient & \(0.005,\,0.01\) & \(0.005\) \\
Discount factor \(\gamma\) & \(0.995,\,0.999,\,1\) & \(0.999\) \\
GAE parameter \(\lambda\) & \(0.95,\,0.97,\,0.99\) & \(0.97\) \\
PPO minibatches & \(4,\,8\) & \(8\) \\
Rollout length per environment & \(256,\,512\) & \(256\) \\
Transformer memory length & \(64,\,128\) & \(64\) \\
Learning-progress exponent \(\rho\) & \(1/2,\,2/3,\,1\) & \(1/2\) \\
Retained observations for LP estimation \(N_{\mathrm{obs}}\) & \(5,\,7\) & \(7\) \\
Uniform-sampling weight in ALP & \(0.2,\,0.3\) & \(0.2\) \\
\bottomrule
\end{tabular}
\end{table}

Most of the evaluated hyperparameter changes had a modest effect on performance. Changes with more substantial negative effects included increasing the entropy coefficient to \(0.01\), extending Transformer memory to \(128\) steps, reducing the learning rate to \(3\times10^{-5}\), and setting the discount factor to \(\gamma=1\). For the discount factor, \(\gamma=0.999\) yielded late-training success rates close to those obtained with \(\gamma=0.995\). We preferred \(\gamma=0.999\) because it discounts delayed goal achievement less heavily, better aligning training with the undiscounted goal-success objective.

\subsection{Sensitivity analysis}
\label{app:sensitivity-analysis}

\begin{figure}[htbp]
    \centering
    \includegraphics[width=\linewidth]{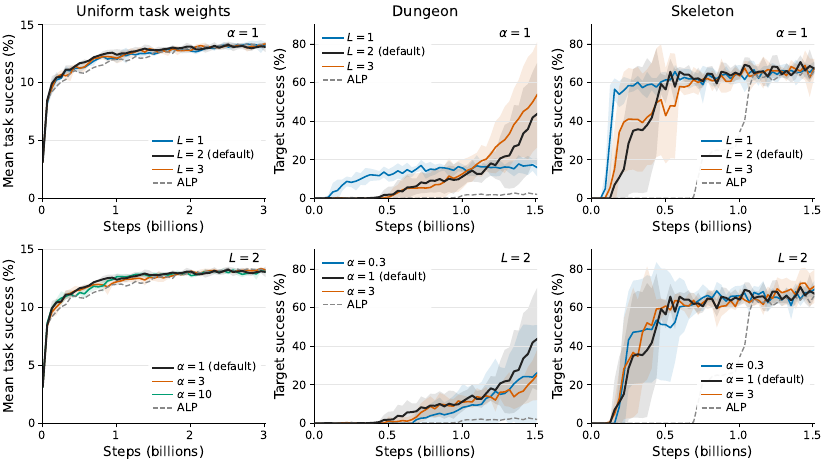}
    \caption{Sensitivity of ALP+D+F to \(L\) (top, \(\alpha=1\)) and \(\alpha\) (bottom, \(L=2\)). Curves show mean evaluation success; shading indicates one sample standard deviation across three seeds under uniform weights and five seeds for each target task.}
    \label{fig:sensitivity-analysis}
\end{figure}

We examine the sensitivity of ALP+D+F to the maximum path length \(L\) used to compute downstream potential and the downstream coefficient \(\alpha\). We consider uniform task weights and settings with all objective weight assigned to either Dungeon or Skeleton. Using the policy restricted to primitive actions, we vary one parameter at a time around the main configuration, keeping all other settings as in the corresponding main experiments. We conducted this analysis after fixing the hyperparameters used in the main experiments; its results did not inform their selection. All configurations and the ALP reference use the same training seeds within each setting: three under uniform weights and five per target. Evaluations follow the protocol in Appendix~\ref{app:evaluation-procedure}. Table~\ref{tab:sensitivity-analysis} reports success at selected training budgets, and Figure~\ref{fig:sensitivity-analysis} shows the learning curves.

The effects of the two parameters depend on the setting and training budget. Under uniform task weights, performance varies little across the tested configurations; on Skeleton, differences mainly concern learning speed, with similar final mean success. Dungeon shows stronger sensitivity: \(L=1\) produces higher mean success early in training but subsequently levels off below \(L=2\) and \(L=3\), while both alternative values of \(\alpha\) yield lower final mean success than the default. These Dungeon comparisons also exhibit substantial variation across seeds. Despite this sensitivity, all tested configurations have higher mean success than ALP at the first reported budget on Skeleton and at the final budget on Dungeon.

\begin{table}[htbp]
\centering
\caption{Sensitivity of ALP+D+F to maximum path length $L$ and downstream coefficient $\alpha$. Entries are evaluation success percentages (mean $\pm$ one sample standard deviation across seeds), averaged over all tasks under uniform weights and measured on the target otherwise. Columns show one-third, two-thirds, and the full budget of $3{,}019{,}898{,}880$ steps under uniform weights or $1{,}509{,}949{,}440$ steps for the targets.}
\label{tab:sensitivity-analysis}
\small
\renewcommand{\arraystretch}{1.08}
\begin{tabular}{@{}lccccc@{}}
\toprule
\multicolumn{6}{@{}l}{\textbf{Uniform task weights} (3 seeds)} \\
 & $L$ & $\alpha$ & $\approx 1.0$B & $\approx 2.0$B & $\approx 3.0$B \\
\cmidrule(lr){4-6}
Default & 2 & 1 & $12.40 \pm 0.49$ & $13.10 \pm 0.19$ & $13.08 \pm 0.46$ \\
 & 1 & 1 & $12.01 \pm 0.33$ & $12.72 \pm 0.12$ & $13.27 \pm 0.21$ \\
 & 3 & 1 & $12.10 \pm 0.45$ & $12.78 \pm 0.17$ & $13.34 \pm 0.06$ \\
 & 2 & 3 & $12.23 \pm 0.07$ & $13.13 \pm 0.12$ & $13.25 \pm 0.11$ \\
 & 2 & 10 & $12.66 \pm 0.29$ & $12.75 \pm 0.18$ & $13.09 \pm 0.13$ \\
\midrule
\addlinespace[3pt]
\multicolumn{6}{@{}l}{\textbf{Dungeon} (5 seeds)} \\
 & $L$ & $\alpha$ & $\approx 0.5$B & $\approx 1.0$B & $\approx 1.5$B \\
\cmidrule(lr){4-6}
Default & 2 & 1 & $1.88 \pm 4.19$ & $11.41 \pm 5.02$ & $43.75 \pm 26.53$ \\
 & 1 & 1 & $11.88 \pm 3.24$ & $15.62 \pm 1.99$ & $15.94 \pm 4.64$ \\
 & 3 & 1 & $0.62 \pm 1.40$ & $12.81 \pm 7.40$ & $53.75 \pm 27.33$ \\
 & 2 & 0.3 & $0.00 \pm 0.00$ & $8.28 \pm 12.30$ & $26.25 \pm 24.67$ \\
 & 2 & 3 & $1.09 \pm 2.45$ & $10.94 \pm 3.54$ & $25.16 \pm 13.16$ \\
\midrule
\addlinespace[3pt]
\multicolumn{6}{@{}l}{\textbf{Skeleton} (5 seeds)} \\
 & $L$ & $\alpha$ & $\approx 0.5$B & $\approx 1.0$B & $\approx 1.5$B \\
\cmidrule(lr){4-6}
Default & 2 & 1 & $56.41 \pm 7.15$ & $64.06 \pm 1.99$ & $67.19 \pm 2.65$ \\
 & 1 & 1 & $61.41 \pm 4.23$ & $64.53 \pm 4.54$ & $67.66 \pm 3.15$ \\
 & 3 & 1 & $42.81 \pm 24.76$ & $63.44 \pm 2.30$ & $67.50 \pm 4.71$ \\
 & 2 & 0.3 & $51.09 \pm 28.97$ & $66.09 \pm 5.08$ & $69.38 \pm 4.54$ \\
 & 2 & 3 & $60.78 \pm 3.38$ & $65.78 \pm 4.11$ & $70.94 \pm 8.58$ \\
\bottomrule
\end{tabular}
\end{table}

\clearpage

\section{Baselines and ablations}
\label{app:baselines-ablations}

Within each comparison, all methods use the same learner and training protocol. Performance and learning-progress estimates are computed as described in Appendix~\ref{app:performance-lp-estimation}. Adaptive curricula use uniform sampling for the first 16 curriculum iterations. In the single-target experiments, each adaptive distribution governs the 60 task selections remaining after the four fixed target selections, as described in Section~\ref{sec:Experiments}.

For compactness, we define the sampling distribution induced by a nonnegative task score \(S\) as
\[
\mathcal N[S](m)
=
\begin{cases}
\displaystyle
\frac{S(m)}{\sum_{m'\in\mathcal T}S(m')},
& \displaystyle\sum_{m'\in\mathcal T}S(m')>0,\\[3mm]
q^{\mathrm U}(m),
& \text{otherwise},
\end{cases}
\]
where \(q^{\mathrm U}(m)=1/|\mathcal T|\) is uniform over the task set.

\subsection{Baselines}
\label{app:baselines}

\paragraph{Uniform sampling (UNIF).}
UNIF samples uniformly over the task set, with \(q_i=q^{\mathrm U}\).

\paragraph{Absolute learning progress (ALP).}
ALP is representative of conventional curricula guided by estimated absolute learning progress \citep{matiisen2019teacher}. It also serves as an ablation of our method with both downstream potential and feasibility-guided sampling removed. Its score and sampling distribution are
\[
\begin{aligned}
S_i^{\mathrm{ALP}}(m)
&=
\left|\widehat{\operatorname{LP}}_i(m)\right|^\rho\,w(m),
\\
q_i
&=
0.8\,\mathcal N[S_i^{\mathrm{ALP}}]
+
0.2\,q^{\mathrm U},
\end{aligned}
\]
with \(\rho=\tfrac12\), as in all LP-based variants.

\paragraph{Intermediate difficulty (I-DIFF).}
I-DIFF represents curricula that prioritize tasks of intermediate difficulty relative to the learner's current competence. We use success-rate thresholds of \(0.1\) and \(0.9\), matching those used by \citet{florensa2018automatic}. The score and sampling distribution are
\[
\begin{aligned}
S_i^{\mathrm{I\text{-}DIFF}}(m)
&=
\mathbf 1\!\left[
0.1\leq\widehat P_i(m)\leq0.9
\right]w(m),
\\
q_i
&=
0.8\,\mathcal N[S_i^{\mathrm{I\text{-}DIFF}}]
+
0.2\,q^{\mathrm U}.
\end{aligned}
\]

\paragraph{Direct target training (TARGET).}
In the single-target experiments, TARGET assigns the entire training budget to the designated target from the first curriculum iteration. It provides a reference for assessing the benefit of training on auxiliary tasks.

\subsection{Ablations}
\label{app:ablations}

\paragraph{Removing feasibility-guided sampling (ALP+D).}
ALP+D retains the full method's LP-based score and omits the feasibility-guided component:
\[
\begin{aligned}
S_i^{\mathrm{ALP+D}}(m)
&=
\left|\widehat{\operatorname{LP}}_i(m)\right|^\rho
\left[
\widetilde w_i^{\mathrm{LP}}(m)
+
\alpha\,\widetilde D_i^{\mathrm{LP}}(m)
\right],
\\
q_i
&=
0.8\,\mathcal N[S_i^{\mathrm{ALP+D}}]
+
0.2\,q^{\mathrm U}.
\end{aligned}
\]
We use \(\alpha=1\) and the normalization defined in Appendix~\ref{app:score-normalization}.

\paragraph{Removing downstream potential (ALP+F).}
ALP+F removes downstream potential from both scores by setting \(\alpha=0\), while retaining feasibility-guided sampling:
\[
\begin{aligned}
S_i^{f,0}(m)
&=
\mathbf 1[\widehat P_i(m)\leq\tau_f]\,
f_i(m)\,w(m),
\\
q_i
&=
0.8\,\mathcal N[S_i^{\mathrm{ALP}}]
+
0.1\,\mathcal N[S_i^{f,0}]
+
0.1\,q^{\mathrm U},
\end{aligned}
\]
with \(\tau_f=0.05\).

\paragraph{Permuting the learning-connectivity graph (PERM).}
PERM follows ALP+D+F but replaces the learning-connectivity weights \(C(s,t)\) with \(C(\sigma(s),\sigma(t))\), where \(\sigma\) is a random permutation of \(\mathcal T\), sampled once per run and held fixed throughout training. Applying the same permutation to both endpoints preserves the graph's topology and edge weights while randomizing its correspondence to the goals. Task weights and the feasibility-evidence graph retain their original goal assignments; all other components and hyperparameters are unchanged.

\section{Additional results and analyses}
\label{app:additional-results}

\FloatBarrier
\subsection{Curriculum dynamics}
\label{app:curriculum-dynamics}

\begin{figure}[htbp]
    \centering
    \includegraphics[width=0.755\linewidth]{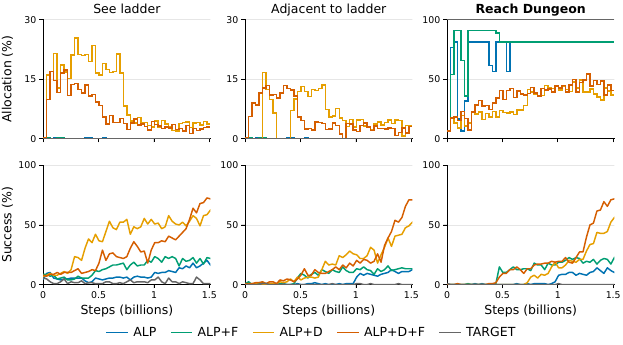}
    \caption{Illustrative curriculum dynamics with all objective weight assigned to reaching the Dungeon. The top row shows the probability mass allocated to the target and selected related auxiliary goals, averaged over non-overlapping blocks of 15 curriculum decisions. The bottom row shows the policy's success rate when evaluated on each goal. For each method, we show the run with the highest final target success among the five seeds in Figure~\ref{fig:target-learning-curves}.}
    \label{fig:target-curriculum-dungeon}
\end{figure}

\begin{figure}[htbp]
    \centering
    \includegraphics[width=\linewidth]{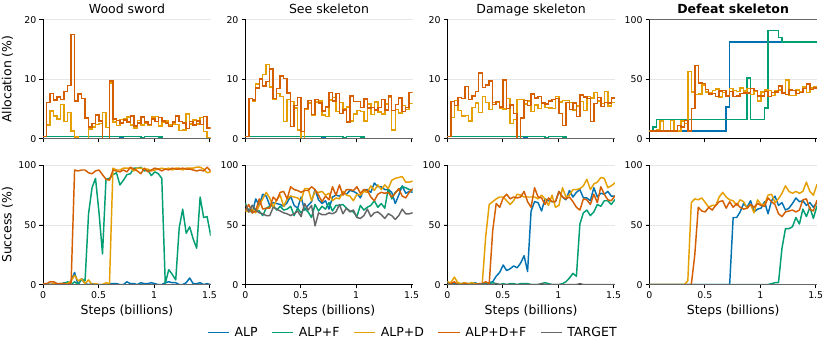}
    \caption{Illustrative curriculum dynamics with all objective weight assigned to defeating a skeleton. The top row shows the probability mass allocated to the target and selected related auxiliary goals, averaged over non-overlapping blocks of 15 curriculum decisions. The bottom row shows the policy's success rate when evaluated on each goal. For each method, we show the run with the highest final target success among the five seeds in Figure~\ref{fig:target-learning-curves}.}
    \label{fig:target-curriculum-skeleton}
\end{figure}

\begin{figure}[htbp]
    \centering
    \includegraphics[width=\linewidth]{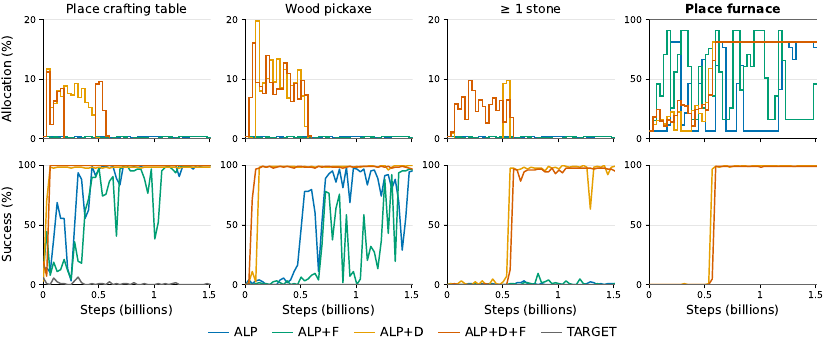}
    \caption{Illustrative curriculum dynamics with all objective weight assigned to placing a furnace. The top row shows the probability mass allocated to the target and selected related auxiliary goals, averaged over non-overlapping blocks of 15 curriculum decisions. The bottom row shows the policy's success rate when evaluated on each goal. For each method, we show the run with the highest final target success among the five seeds in Figure~\ref{fig:target-learning-curves}.}
    \label{fig:target-curriculum-furnace}
\end{figure}

\paragraph{Illustrative trajectories across methods.}
Figures~\ref{fig:target-curriculum-dungeon}--\ref{fig:target-curriculum-furnace} extend the Wood example in Figure~\ref{fig:target-curriculum-wood} to Dungeon, Skeleton, and Furnace, using each method's run with the highest final target success.

\begin{figure}[htbp]
    \centering
    \includegraphics[width=\linewidth]{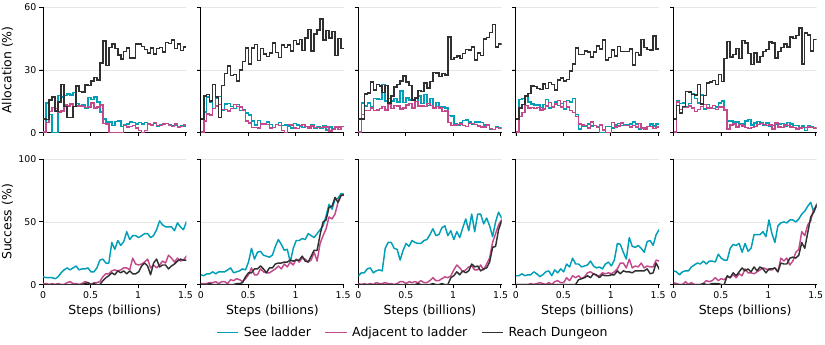}
    \caption{Curriculum dynamics of ALP+D+F with all objective weight assigned to reaching the Dungeon. Each column shows one of the five training seeds in Figure~\ref{fig:target-learning-curves}. The top row shows allocation probability averaged over non-overlapping blocks of 15 curriculum decisions; the bottom row shows evaluation success.}
    \label{fig:target-curriculum-dungeon-all-seeds}
\end{figure}

\begin{figure}[htbp]
    \centering
    \includegraphics[width=\linewidth]{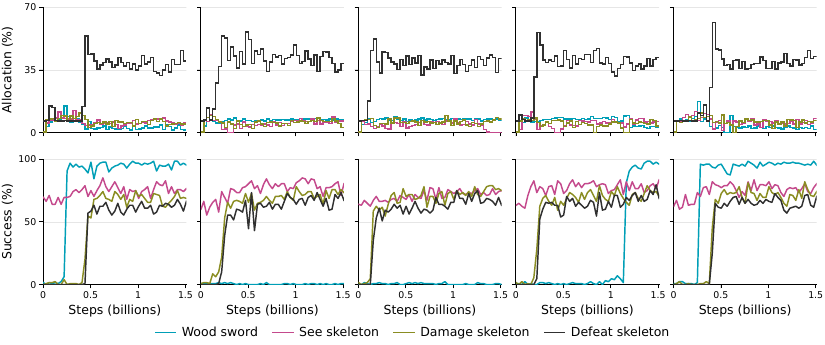}
    \caption{Curriculum dynamics of ALP+D+F with all objective weight assigned to defeating a skeleton. Each column shows one of the five training seeds in Figure~\ref{fig:target-learning-curves}. The top row shows allocation probability averaged over non-overlapping blocks of 15 curriculum decisions; the bottom row shows evaluation success.}
    \label{fig:target-curriculum-skeleton-all-seeds}
\end{figure}

\begin{figure}[htbp]
    \centering
    \includegraphics[width=\linewidth]{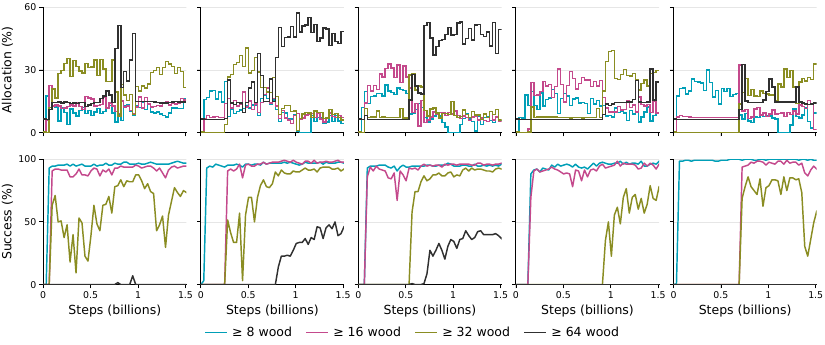}
    \caption{Curriculum dynamics of ALP+D+F with all objective weight assigned to having at least 64 wood. Each column shows one of the five training seeds in Figure~\ref{fig:target-learning-curves}. The top row shows allocation probability averaged over non-overlapping blocks of 15 curriculum decisions; the bottom row shows evaluation success.}
    \label{fig:target-curriculum-wood-all-seeds}
\end{figure}

\begin{figure}[htbp]
    \centering
    \includegraphics[width=\linewidth]{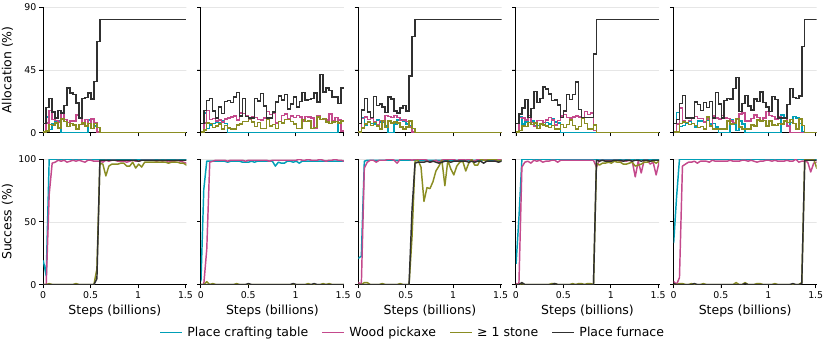}
    \caption{Curriculum dynamics of ALP+D+F with all objective weight assigned to placing a furnace. Each column shows one of the five training seeds in Figure~\ref{fig:target-learning-curves}. The top row shows allocation probability averaged over non-overlapping blocks of 15 curriculum decisions; the bottom row shows evaluation success.}
    \label{fig:target-curriculum-furnace-all-seeds}
\end{figure}

\FloatBarrier
\paragraph{ALP+D+F across training seeds.}
Figures~\ref{fig:target-curriculum-dungeon-all-seeds}--\ref{fig:target-curriculum-furnace-all-seeds} show individual ALP+D+F trajectories for all five training seeds on each target.

\FloatBarrier
\subsection{Smoothing the target-centered objectives}
\label{app:target-weight-smoothing}

\begin{figure}[htbp]
    \centering
    \includegraphics[width=\linewidth]{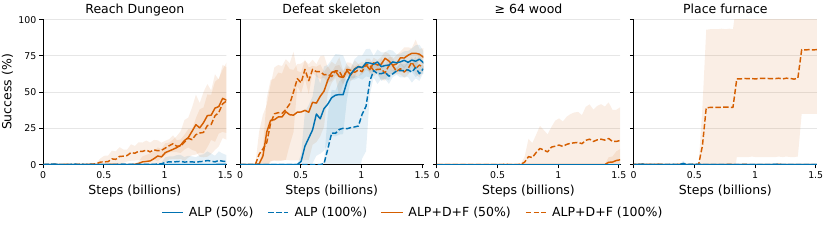}
    \caption{Evaluation success on selected target tasks during training, with 50\% (solid) or 100\% (dashed) of the objective weight assigned to the corresponding target. Lines show means across five seeds; shaded regions indicate one standard deviation.}
    \label{fig:target-weight-comparison}
\end{figure}

With all objective weight assigned to the target, ALP cannot prioritize auxiliary tasks according to their learning progress. To test whether the benefits of ALP+D+F discussed in Section~\ref{sec:results-target-tasks} persist when this restriction is removed, we evaluate ALP with 50\% of the total objective weight assigned to the target and the remaining 50\% distributed uniformly across the other tasks (Figure~\ref{fig:target-weight-comparison}). Reweighting accelerates skeleton learning under ALP, but learning remains slower than under ALP+D+F. Moreover, ALP's Dungeon performance declines, and it remains unsuccessful on Wood and Furnace. Applying the same reweighting to ALP+D+F has its most pronounced effects on Wood and Furnace, where success rates are substantially reduced within the tested budget.

\subsection{Discussion of downstream potential}
\label{app:downstream-feasibility-discussion}

\paragraph{Downstream potential aggregation.}
The strongest-path construction is a heuristic for aggregating LLM judgments into a measure of potential downstream usefulness. It combines contributions from distinct downstream tasks according to their importance, retaining only the strongest discounted path to each task to avoid counting alternative routes separately. Multiplying edge weights and applying the path discount attenuates weakly supported connections and longer chains. These choices provide a simple aggregation rule, without assigning the resulting path scores a calibrated interpretation as transfer probabilities or performance gains. A single learning-connectivity edge can represent potential for either immediate transfer or more productive subsequent training on another task. Consequently, \(L=1\) already allows effects on future learning to influence downstream potential. Increasing \(L\) extends the aggregation to possible effects through intermediate tasks; it limits the number of composed relationships without specifying how many subsequent training decisions their realization may require. The aggregation does not explicitly model the intermediate learning required to realize the opportunities suggested by these paths.

\end{document}